\documentclass[final,3p,natbib]{elsarticle}

\usepackage{hyperref}
\usepackage{amsfonts}
\usepackage{gensymb}
\usepackage{subfigure}
\usepackage{tabularx}
\usepackage{bm}
\usepackage{xcolor}
\usepackage{amsmath}
\usepackage{algorithmic}
\usepackage[linesnumbered,ruled]{algorithm2e}
\usepackage{graphicx}
\usepackage{makecell}

\begin{document}

\begin{frontmatter}

\title{Deep multi-prototype capsule networks}



\author[First]{Saeid Abbassi}
\ead{s.abbaasi@mail.um.ac.ir}

\author[First]{Kamaledin Ghiasi-Shirazi}
\ead{k.ghiasi@um.ac.ir}

\author[First]{Ahad Harati}
\ead{a.harati@um.ac.ir}

\address[First]{Department of Computer Engineering, Ferdowsi University of Mashhad, Mashhad, Iran}

\begin{abstract}
Capsule networks are a type of neural network that identify image parts and form the instantiation parameters of a whole hierarchically. The goal behind the network is to perform an inverse computer graphics task, and the network parameters are the mapping weights that transform parts into a whole. The trainability of capsule networks in complex data with high intra-class or intra-part variation is challenging. This paper presents a multi-prototype architecture for guiding capsule networks to represent the variations in the image parts. To this end, instead of considering a single capsule for each class and part, the proposed method employs several capsules (co-group capsules), capturing multiple prototypes of an object. In the final layer, co-group capsules compete, and their soft output is considered the target for a competitive cross-entropy loss.
Moreover, in the middle layers, the most active capsules map to the next layer with a shared weight among the co-groups. Consequently, due to the reduction in parameters, implicit weight-sharing makes it possible to have more deep capsule network layers. The experimental results on MNIST, SVHN, C-Cube, CEDAR, MCYT, and UTSig datasets reveal that the proposed model outperforms others regarding image classification accuracy.
\end{abstract}


\end{frontmatter}


\section{Introduction}
\label{sec:introduction}
The applications of computer vision have become widespread in the past decade in different tasks, including medicine \cite{dougherty2020digital}, intelligent transportation \cite{chang2020ai}, and authentication \cite{dubey2019review}. As applications grow, learning algorithms need to be associated with more complex datasets than before. The structure of brain neurons inspires neural networks, which categorize into several varieties, including perceptrons, convolutional neural networks, and recurrent networks. In these networks, the corresponding part neurons are activated when a part (such as an eye) exists. Activating a sufficient number of class parts activates a class neuron (such as a face) in the output layer. However, the image-part relations are not generally modeled \cite{hinton2011transforming}.

Hinton et al. \cite{hinton2011transforming} introduced an auto-encoder network that comprises transformation involving a general representation of images. As a result, they defined a capsule as a specific number of neurons that not only capture the presence but also model some information about a part, such as size, orientation, and texture. Using a part-to-whole hierarchy, capsule networks outperform traditional learning models in some cases.

Subsequently, capsule networks perform hierarchically based on the image part recognition and construction of a whole. The capsule network has lately performed well in image classification \cite{sabour2017dynamic}. The capsule layers perform an inverse computer graphic operation \cite{hinton2018matrix}. Each capsule represents a part, and each part votes for the instantiation parameters of every whole using mapping matrices. For example, if several parts (such as the eyes, nose, and mouth) agree on the pose and attributes of a ``whole'' (such as the face), then the existence possibility of the ``whole'' increases. In \cite{sabour2017dynamic}, the length of the activity vector of each capsule indicates the detection probability of an object. Each capsule predicts the pose of a ``whole'' using a mapping matrix. When various predictions supplied by multiple parts have a consensus, the higher-order capsule is also activated.

In \cite{qian2021image} and \cite{li2021robustness}, the appropriateness and robustness of capsule networks in comparison to convolutional neural networks were studied, particularly in facial expression recognition.

Capsule networks perform well on datasets with no background (such as MNIST). Because of their monochrome structure and lack of background, the experiments show capsule network outperforms standard convolutional models in signature image recognition \cite{jampour2021capsnet}. On the other hand, experiments reveal that the capsule networks perform poorly when the intra-class variation is high \cite{sabour2017dynamic}. It has also been demonstrated in \cite{xi2017capsule} that increasing the capsule layers does not improve performance since the number of parameters grows extremely. For example in the MNIST classification problem, there are about seven million parameters in a standard two-layer capsule network \cite{shi2022sparse}.

Many efforts have been made to increase the performance of capsule networks when dealing with complex data. Since training a capsule network necessitates extensive computing resources, studies have tried optimizing the network training process. A multi-scale capsule network to improve the computational efficiency of the capsule network was proposed in \cite{xiang2018ms}. Instead of using convolution layers to construct the initial capsules like in the standard model, this architecture employs three levels of multi-scale feature extraction. The results of all three units are combined to create the initial capsules. Because the features are scalable, the number of initial capsules can be lowered, thereby reducing network parameters.

For multi-prototyping capsule networks, attempts were made in a simple method in \cite{do2019multi} and extended in \cite{do2021efficiency}. The main purpose of these two studies is to develop a framework for parallelizing the training process of capsule networks. In this regard, multiple capsule layer lanes are parallel to produce specified dimensions of output layer capsules \cite{do2021efficiency}. Although the multi-line model includes some multi-prototyping, it is clear that this network lacks a condition to encourage distinct lines to form intra-class clusters, and the main application of this architecture is the parallelization of heavy processing in the capsule networks.

To address the mentioned issues with the capsule networks in the face of high intra-class variations, we propose a deep multi-prototype capsule network that extends our previous work \cite{abbaasi2023multi}. The method proposed in \cite{abbaasi2023multi} learns the distinct prototypes in the dataset at the whole and part levels simultaneously. Thus, several capsules are employed instead of considering a single capsule for each class in the final layer. As a result, capsules from the same group (co-group capsule) compete with each other during the learning process, and the output of the competition plays a role in the loss function calculations. Moreover, the model is generalized at the part-level capsules. As mentioned earlier, each middle layer capsule represents an image part. Since the image parts have different prototypes that can be pretty diverse, in this paper, we propose to consider multiple prototypes in the middle layer of the capsule network, where each part prototype has a distinct activity depending on the input image. For each part, the most active capsule is sent to the next layer. As we assume a weight-sharing between the co-group capsules of each part, it made it possible to have deeper capsule networks.

The rest of the paper is organized as follows. Section \ref{sec:Related work} presents the related works of the capsule networks and multi-prototyping. The proposed multi-prototype deep capsule network and its architecture are provided in Section \ref{sec:The proposed method}. The experimental results and comparisons of the proposed model with similar methods are in Section \ref{sec:Experimental Result}, and the paper concludes in Section \ref{sec:conclusion} with a discussion.

\section{Related work}
\label{sec:Related work}
Deep convolutional networks have received much interest due to their outstanding performance, particularly in computer vision. These networks comprise various layers such as convolution, max-pooling, batch normalization, and fully connected layers. The primary convolutional layers extract basic image features such as color, gradients, and edges, while the final layers extract higher-level features using weighted combinations of primaries \cite{zhang2018visual}. Despite their promising performance when enough data is provided for training, a significant limitation is that they ignore the relations between distinct image parts.

It is demonstrated in \cite{hinton2011transforming} that neural networks can learn features like those used in SIFT, which include a general description of visual elements. Hinton et al. proposed the idea of capsules in \cite{hinton2011transforming} and stated it could make superior to conventional models since the part-to-whole recognition and relations between them are viewpoint invariant. Moreover, Hinton et al. proposed a capsule network model \cite{sabour2017dynamic}. The term capsule refers to cover and uses it because this method covers the essential aspects of image parts in a vector. As previously stated, each capsule is a group of neurons whose pose represents the instantiation of a whole or part. For a part, instantiation consists of information about position, size, orientation, texture, etc. In \cite{sabour2017dynamic}, the capsule's activity vector length is used as the probability of the corresponding object being present. Each capsule uses a mapping matrix to predict the pose of a ``whole''. When multiple votes agree on a ``whole'' position, the parent capsule activation also increases. The method of determining agreement location among votes has resulted in the development of various routing algorithms.

Accordingly, the goal of dynamic routing \cite{sabour2017dynamic} is to introduce capsule networks and a type of routing between child and parent layer capsules. This network has two capsule layers, called primary and digit caps, followed by two convolution layers. After the convolution layers, a capsule is formed by a specific number of neurons having the same $x$ and $y$ on different convolution planes. The resulting child capsules are mapped to the parent capsules using transforming matrices. The parent capsules generally have the same number of classes as the dataset. In other words, the $i$-th primary capsule map to the $j$-th parent capsule by matrix $W_{ij} \in \mathbb{R} ^{8\times16}$.Finally, each capsule in the last layer is a compact representation of each class. Assume the output of the $i$-th primary capsule and the input of the $j$-th digit capsule are denoted by $u_i$ and $s_j$, respectively. We have:
\begin{equation}
\label{eq:eq01}
\begin{split}
\hat{u}_{(j|i)} &= W_{ij} u_i \\
s_j &= \sum_i{c_{ij} \hat{u}_{(j|i)}},
\end{split}
\end{equation}
where $c_{ij}$ is called a coupling coefficient, indicating the relation between the $i$-th capsule of the current layer and the $j$-th next layer capsule. In other words, the more compatible the part represented by the $i$-th capsule of the child layer is with the $j$-th capsule of the parent layer, the larger $c_{ij}$ is, and vice versa. Please note that the sum of the coupling coefficients coming out of a capsule equals one. The coupling coefficients are calculated by formula 
\begin{equation}
\label{eq:eq02}
c_{ij}= \frac{\exp{(b_{ij})}}{\sum_k\exp{(b_{ik})}} .
\end{equation}

In dynamic routing, an unsupervised iterative algorithm calculates $b_{ij}$. In this way, the output of the $j$-th parent capsule is calculated by equation \ref{eq:eq03}. Afterward, the correlation between $\textbf{v}_j$ and $\hat{u}_{j|i}$ for the $i$-th child capsule calculated by the inner product adds to $b_{ij}$. Note that in equation \ref{eq:eq03}, a nonlinear function called squashing is applied to the vector $s_j$.
\begin{equation}
\label{eq:eq03}
\textbf{v}_j = \frac{\Vert s_j\Vert ^2}{1+\Vert s_j\Vert^2}\frac{s_j}{\Vert s_j\Vert}
\end{equation}

On the MNIST dataset, dynamic routing achieved 99.75\% accuracy in \cite{sabour2017dynamic}, the highest accuracy recorded. The time-consuming process of locating the primary cluster is one of the dynamic-routing areas for improvement. Additionally, dynamic routing does not work efficiently in the case of image noises. In \cite{hahn2019self}, a simple self-routing method is proposed that assigns the routing process to the network (not as a separate task). Thus, in addition to the mapping matrices, the parameters for routing are also trained \cite{hahn2019self}. Despite the mentioned studies, \cite{paik2019capsule} has a detailed discussion on different routing methods. The conclusion indicates that the current routing algorithms do not affect the recognition rate much, and efficient routing methods are needed.

In EM (Expectation-Maximization) routing \cite{hinton2018matrix}, the authors made three modifications to the dynamic routing model. First, each capsule has a matrix form instead of a vector. Also, unlike dynamic routing, which uses the normalized output length of each capsule as its activity, EM routing uses a separate parameter that maintains the activation probability.

Nevertheless, the main idea is a new routing process that searches for a Gaussian distribution, including consensus, with an approach like the Gaussian mixture model.

In \cite{bahadori2018spectral}, a spectral capsule network was presented by applying modifications to EM-based routing. The spectral capsule networks aim to identify the consensus of the components, much like EM routing \cite{hinton2018matrix}, except that the agreement is performed by evaluating the level of alignment in a linear subspace rather than by finding a dense cluster. Some studies proposed using pre-trained model features as input of the capsule networks to improve the capsule network input features. The capsule network has performed iris classification and identification as an authentication method \cite{zhao2019deep}. This approach uses well-known convolutional networks, including VGG, ResNet, and Inception. Also, in \cite{jampour2021capsnet} and \cite{javidi2021covid}, ResNet and DenseNet are connected to the capsule network, respectively. The resulting architecture improved the performance since the standard model suffers from encapsulating weak features.

Ghiasi-shirazi\citep{ghiasi2019competitive} introduces a competitive cross-entropy loss function that performs classification using a single-layer network for nonlinearity separable datasets like MNIST. In contrast, a single-layer neural network can only address linear problems. This approach considers multiple output neurons for each class rather than just one to capture the intra-class prototypes. In \cite{abbaasi2023multi}, the multi-prototype approach is extended to the capsule networks. Abbaasi et al. \citep{abbaasi2023multi} introduced a capsule network architecture in which multiple capsules are considered in the final layer instead of one capsule for each class. The co-group capsules compete with each other through a competitive cross-entropy loss function.

As stated in \cite{abbaasi2023multi}, instead of using one capsule for each class k in the final layer (layer L) of the capsule network, an arbitrary number $|O_k^L|$ of capsules are used, where $O_k^L$ indicates the set of indices corresponding to the final capsules of class $k$. As a result, the total number of capsules of the last layer is $n = \sum_k|O_k^L|$. It is while there are $c$ capsules (as the number of classes) in the final layer of the standard capsule networks.

For this purpose, first, the output of the $j$-th capsule of the parent layer, indicated by $\textbf{v}_j$, is calculated as follows:
\begin{equation}
\label{eq:eq04}
\textbf{v}_j = \frac{\Vert s_j\Vert ^2}{1+\Vert s_j\Vert^2}\frac{s_j}{\Vert s_j\Vert} \quad \textrm{for} \; 1\leq j \leq n=\sum_k |O_k |.
\end{equation}

Afterward, the probability distribution of the network output is defined as:
\begin{equation}
\label{eq:eq05}
y_j = \frac{\Vert\textbf{v}_j\Vert}{\sum_{t=1}^{n} \Vert\textbf{v}_t\Vert} \quad \textrm{for} \; 1\leq j \leq n.
\end{equation}

We next define the target distribution. Assuming that the input image belongs to class $k$, we set the target value to zero for each output capsule that does not belong to $O_k$. Further, the target probability mass for a capsule $i\in O_k$ is defined as follows:
\begin{equation}
\label{eq:eq06}
 \tau_i = \frac{\Vert\textbf{v}_{O_{k,i}}\Vert}{\sum_{j=1}^{|O_k|} \Vert\textbf{v}_{O_{k,j}}\Vert},
\end{equation}
where $\| \textbf{v}_{O_{k,i}} \|$ is the second norm of the $i$-th capsule of class $k$. Finally, the competitive cross-entropy loss function is defined as follows:
\begin{equation}
\label{eq:eq07}
 E^{CCE}=-\sum_{i=1}^{n} \tau_ilog(y_i).
\end{equation}

It is also proved that the derivative of $E^{CCE}$ with respect to $\textbf{v}_s$ is:
\begin{equation}
\label{eq:eq08}
\frac{\partial{E^{CC
 E}}}{\partial{\textbf{v}_s}}
 \frac{\textbf{v}_s}{\Vert \textbf{v}_s\Vert} \textbf{1}^T \left(\frac{\tau}{\Vert \textbf{v}\Vert} \odot (\textbf{y} - o^s)\right).
\end{equation}
Experimental results of \cite{abbaasi2023multi} demonstrated the promising performance of the multi-prototype capsule network compared to the standard capsule network.

In order to discover different intra-part prototypes besides intra-class ones, in the proposed method, in addition to using multi-prototype in the final layer to represent the intra-class variations, the multi-prototype architecture is also generalized to the part level in middle layers. Also, the features of the DenseNet are connected to the proposed architecture to help the network discover aligned prototype features and improve the quality of the input features.

\section{The proposed method}
\label{sec:The proposed method}

Real-world datasets commonly contain a considerable amount of intra-class variation. These variations mainly originate from the difference in the constituent components of the entity. For example, the mentioned variations depend on not only the variations in human faces but also many variations evident in face parts. The high intra-class and intra-part variations make the recognition operation challenging for the standard capsule networks model.

\begin{figure}[htp]				
	\centering
	\includegraphics[scale=.4]{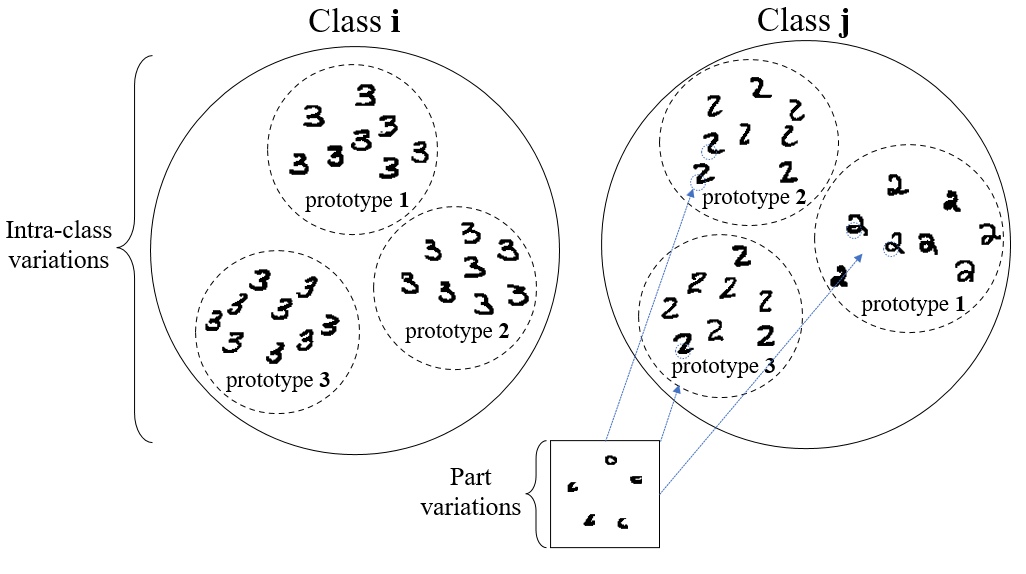}
	\caption{Examples of digits 2 and 3 in the MNIST dataset. The prototype variations for each class as well as each part, are indicated in this figure.}
	\label{fig:figure 1}
\end{figure}

As mentioned, the base capsule network is unsuccessful in dealing with complex data. This is while the capsule networks benefit from a rich learning capacity due to the high number of parameters. The complexity of real-world data comes from several different sources, such as displacement, rotation, and scale. Another source of data complexity is the prototype variations in the parts of an entity, as well as intra-class variation. Consider Figure \ref{fig:figure 1}, illustrating a simple two-class classification problem. In this figure, the intra-class variation can be seen for each class.
On the other hand, the diversity also spreads to the image parts. An example of the part variety related to the bottom-left corner of number 2 is depicted in Figure \ref{fig:figure 1}. In the base capsule networks, there is a part-to-whole process, and each child capsule represents a part in the image, but the assumption of the part variety is not considered in the training process.

We introduce a model to train a capsule network facing complex data with intra-class and intra-part variations. The proposed method also provides an approach to make the capsule networks deep, which improves the performance while keeping the network trainable due to the implicit weight-sharing between multiple prototypes of each part. However, the increased number of capsule layers disturbed its trainability before \cite{xi2017capsule}.

In the middle layers, each capsule identifies an image part. Since the image parts may have different prototypes, which sometimes have much diversity, in the middle layer of the network, we consider several prototypes for each part. Depending on the input image, each will have different norm activity, and the most active capsule among the co-groups is sent to the next layer using a shared transforming weight. In the final layer, each capsule represents the instantiation parameters of a whole (class). In this case, having multiple capsules per class (instead of one capsule) makes the network more robust to the whole variations.

\begin{figure}[htp]				
	\centering
	\includegraphics[scale=.51]{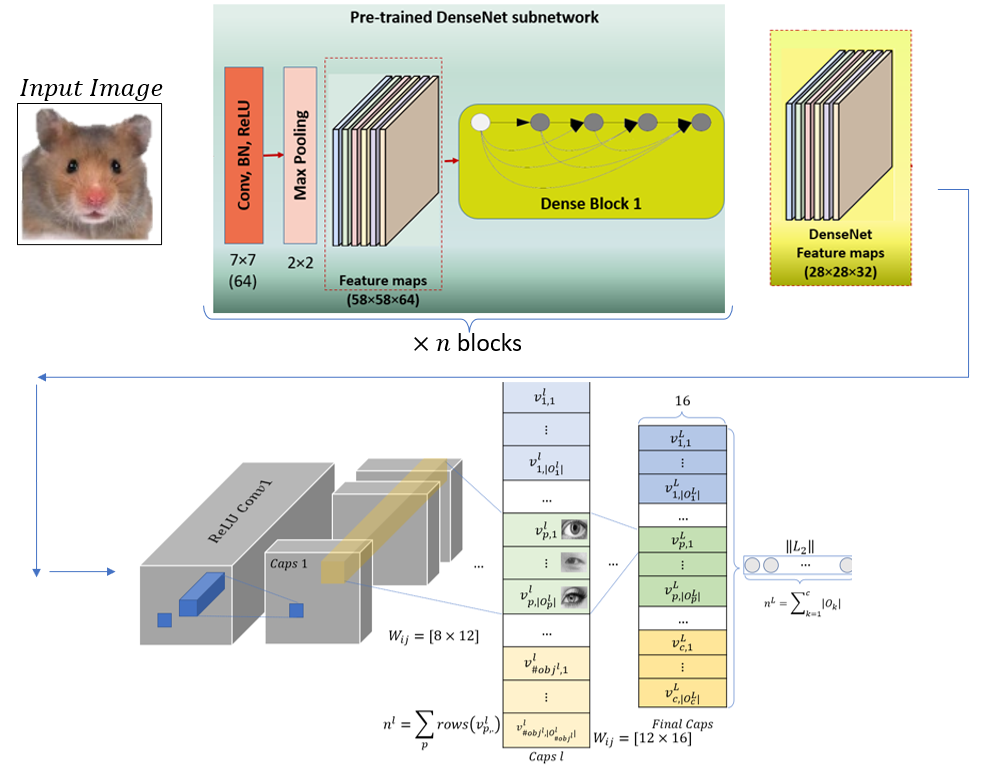}
	\caption{Hybrid DenseNet and multi-prototype deep capsule network architecture. In this architecture, the input of the proposed capsule network is supplied by the DenseNet features.}
	\label{fig:figure 2}
\end{figure}

As shown in Figure \ref{fig:figure 2}, the number of capsule layers in the proposed architecture has increased to several instead of two in the standard model. Increasing the number of layers causes a disturbance in the learning process of the standard capsule networks due to a high volume of transforming weight matrices. Besides, the multiplicity of capsule layers is essential due to the hierarchical nature of part-to-whole recognition. In the proposed method, by considering multiple prototypes at the part level, a deep architecture is provided for the capsule networks, bringing better performance and making learning more accessible.

Suppose $v^l$  is a matrix containing $n^l$ capsules in the $l$-th layer of the capsule network. Also, $v_{p,.}^l$ contains all capsule prototypes describing the $p$-th part (for example, the eye) in the $l$-th layer. Finally, consider $v_{p,k}^l$ the capsule describing the $k$-th prototype of the $p$-th part in the $l$-th layer. For a part p, there are $|O_p^l|$ number of capsules in the $l$-th layer where the vector $O_p^l$ holds the indices of the corresponding capsules. Hence, the total number of capsules in the $l$-th layer is:
\begin{equation}
\label{eq:eq09}
n^l = \sum_p |O_p^l|
\end{equation}

In the phase of mapping the capsules of the $l$-th child layer to the $(l+1)$-{th} parent layer, the most active capsule (with the largest vector length) from $v_{p,.}^l$ is selected as the output capsule from its group. The less active capsules that lose the competition are masked or filtered out. Due to implicit weight-sharing, using this approach, the number of mapping matrices for capsules between the $l$-th layer and the $(l+1)$-th layer is $S^l\times n^{l+1}$. The parameter $S^l$ represents the number of unique parts in the $l$-th layer. Thus, the weight-sharing helps to reduce the parameters and subsequently increase the trainability of the network in the condition that the number of capsule layers grows.

The co-group capsules should be guaranteed to have a common receptive field on the input image where a specific part exists. In the layer before the primary capsules, i.e., the second convolution layer, the neurons located on a common $x$ and $y$ have the same receptive field of the image, and the co-group capsules are selected among them. Assuming the depth of the convolution layer is $D$, in a common $x$ and $y$, there are $D/d_1$ capsules, where $d_1$ indicates the dimensions of a primary capsule. By dividing $D/d_1$ fairly into classes, the number of capsules of each class is determined to be categorized into co-groups. Also, the number of prototypes in the subsequent capsule layers is chosen arbitrarily. In the final layer, the number of prototypes indicates the intra-class variation. In this layer, instead of selecting only one capsule (the most active) among the co-groups, a soft competition is held among the prototypes of desired class for loss function calculations as in \cite{abbaasi2023multi}.

The DenseNet model \cite{he2016deep} is a successful deep neural network that has achieved good results in various tasks in recent years. It is an extended architecture of ResNet \cite{huang2017densely}. Unlike ResNet, which uses shortcut connections to transfer input data to subsequent blocks, DenseNet exploits the effect of shortcut connections for subsequent blocks. This operation is such that the input image is transferred to the subsequent blocks, and the result of the previous blocks is similarly sent to the following blocks. The architecture offers valuable advantages such as reducing vanishing gradients (like ResNet), encouraging feature reuse, and enhancing feature propagation. Also, at the same time, the number of parameters of this network has been effectively reduced.

Since the feature extraction process in the capsule networks is not rich enough, in the proposed model, DenseNet is used as an efficient feature extraction, and the features of its middle blocks are connected to the input of the multi-prototype capsule network. As illustrated in Figure \ref{fig:figure 2}, in the proposed model, we first train DenseNet and then extract the features of the second block. These features are considered neither low-level nor high-level features but contain desirable information. Especially for datasets with non-aligned objects where image parts may not have a shared location across all images, DenseNet features make the co-group capsules robust against unintentionally involving a stranger part capsule into the group of a particular part.

Inspired by \cite{jampour2021capsnet} and \cite{javidi2021covid}, in the proposed method, we use the following regularization term for capsule networks, which removes the reconstruction sub-network in \cite{sabour2017dynamic}. The weight decay Frobenius norm regularization helps to prevent network complexity and reduce the parameters. Therefore, the final loss function is defined as follows:

\begin{equation}
\label{eq:eq10}
L_{Final} = L_{Total}+\frac{\beta}{\sigma\sqrt{n_{L-1}\times n_L \times d_{L-1} \times d_L }} \| W \|_F \,
\end{equation}

The hyper-parameter $\beta$ indicates the importance of the regularization term and is a positive  real number close to zero. Also, we provide a normalization coefficient for $\|W\|$ in the denominator, which balances the two parts of the loss function. Suppose the network contains $L$ capsule layers, and the number of capsules in the $l$-th layer is $n_l$. Also, the dimension of the $l$-th capsule layer is $d_l$. Considering the tensor $W$ contains all the mapping matrices between all the pairs of capsules of layer $L$ and $L-1$, where each includes a mapping matrix of size $d_{L-1}\times d_L$, the overall size of the $W$ is 
$size(W)=n_{L-1} \times n_L \times d_{L-1} \times d_L$.
We assume that each entry of $W$ is initialized by sampling from a zero-mean normal distribution with standard deviation $\sigma$, i.e., $W_{i,j}\sim N(0,\sigma)$ for all valid $i$ and $j$. According to the following, the square of each of its elements has an expected value equal to $\sigma^2$. $E\{W \odot W\} \rightarrow \sigma^2$.
Therefore:
\begin{equation}
\label{eq:eq11}
\| W \|_F \approx \sqrt{size(W) \times \sigma^2} = \sigma \sqrt{n_{L-1} \times n_L \times d_{L-1} \times d_L }
\end{equation}

\section{Experimental results}
\label{sec:Experimental Result}
This section evaluates the proposed multi-prototype deep capsule model that uses DenseNet features as input. We use four capsule layers in the experiments, and in all layers (except the final layer), the output of only the winner capsule between prototypes is transferred to the next layer. For the last layer, we follow the instructions in \cite{abbaasi2023multi}.

\subsection{Datasets}
In order to evaluate the proposed model, several image classification datasets have been used. The MNIST dataset includes handwritten numbers, the SVHN dataset contains images of house license plate numbers from the street view, the C-Cube dataset contains cursive images of lower and uppercase handwritten letters of the English alphabet, and the CEDAR \cite{kalera2004offline}, and MCYT \cite{diaz2019perspective} datasets contain images of human signatures. UTSig dataset  \cite{calik2019large} also contains 8280 images from 115 signers. We provided the details of the datasets in Table \ref{tab:table 1}.

\begin{table}[h]				
	\caption{Specification of each dataset. In the last column, the number of training and test images are separated by.}	
	\label{tab:table 1}			
	\centering
	\begin{tabular}{ l c c c c }
	    \Xhline{2\arrayrulewidth}
	    dataset	& \#images & image size & \#Channels & \#Class\\
		\hline
		MNIST & 60000/10000	& 28x28 & 1 & 10\\
		SVHN & 32257/26032 & 32x32 & 3 & 10\\
		C-Cube & 32160/19133 & 32x32 & 1 & 52\\
		CEDAR & 880/440 & 64x64 & 1 & 55\\
		MCYT & 825/300 & 64x64 & 1 & 75\\
            UTSig & 6210/2070 & 64x64 & 1 & 115\\
		\Xhline{2\arrayrulewidth}
	\end{tabular}				
\end{table}

\subsection{Settings}
In the experiments, we use Adam optimizer with a learning rate of $0.001$, $\beta_1=0.9$, $\beta_2=0.999$, and $\varepsilon=10^{-8}$. The total number of epochs is $100$. For the first and second convolution layers, filters are set to $256$ with ReLU activation function and strides equal to one and two, respectively. Due to the memory limitation related to graphics hardware, parameters such as the number of initial capsules and competitive prototypes differ for each dataset.

\subsection{Evaluations}
The evaluation of the proposed method compared to \cite{sabour2017dynamic} on the mentioned datasets is provided in Table \ref{tab:table 2}. The results show the superiority of the proposed model in all datasets compared to the standard capsule network.

\begin{table}[h]				
	\caption{The classification accuracy results of the proposed method compared to the dynamic routing \cite{sabour2017dynamic}}
	\label{tab:table 2}			
	\centering
	\begin{tabular}{ l c c}
	    
	    \Xhline{2\arrayrulewidth}

	     Dataset & \multicolumn{2}{c}{Accuracy (\%)}\\
	    \cline{2-3}
	    	& Dynamic routing &The proposed model\\
		\hline
		MNIST &	99.29  & \textbf{99.56}\\
		SVHN & 89.77  & \textbf{91.83}\\
		C-Cube & 80.44  & \textbf{83.96}\\
	    CEDAR & 95.45 & \textbf{100}\\
		MCYT & 87.00  & \textbf{99.66}\\
            UTSig & 87.58  & \textbf{99.38}\\
		\Xhline{2\arrayrulewidth}

	\end{tabular}				
\end{table}

Experiments have also been conducted in various training and test data division conditions in signature datasets. We compared the evaluation results with the standard capsule network, DenseNet \cite{he2016deep}, a standard capsule network equipped with the regularization term (without multi-prototyping). Then the results were compared to the state-of-the-art models that have obtained promising results in the field of signature image recognition.

Due to different conditions for using data in signature identification datasets among related works, we adopt multiple protocols to evaluate the proposed method compared to the others. Similar to \cite{jampour2021capsnet}, three protocols are used for evaluation, and five training and test subsets are randomly generated for each protocol. In the first protocol, we use $25\%$ of the data for training and $75\%$ for testing. In the second protocol, we use $50\%$ of the data for training and $50\%$ for testing. Finally, we use $75\%$ of the data for training in the third protocol and the remaining $25\%$ of the data for testing. We use such evaluations to measure the capability of the proposed method in the presence of a small amount of training data.

Additionally, signature images of the same size are used in all experiments. For all images, preprocessing was used to center the signatures. Finally, images were resized to $200\times200$, followed by thresholding to make them binary.

In the comparisons, the standard DenseNet, the standard capsule network, the capsule network equipped with the regularization term (without using multi-prototyping), and finally, our multi-prototype architecture are present.

In Table \ref{tab:table 3}, Table \ref{tab:table 4}, and Table \ref{tab:table 5}, the signature recognition accuracy of the proposed method is shown in comparison with other methods. As can be seen, the proposed method, i.e., GMP-CapsNet, obtained the best results in all experiments on all three datasets.

\begin{table}[h]
	\caption{Evaluation results of different methods on the CEDAR dataset with three protocols 25-75, 50-50, and 25-75\% ratio of training and test data for each signer}	
	\label{tab:table 3}				
	\centering
	\begin{tabular}{ l c c c}
	    
	    \Xhline{2\arrayrulewidth}

	     Methods & \multicolumn{3}{c}{CEDAR}\\
	    \cline{2-4}
	    	& 25-75 & 50-50 & 75-25\\
		\hline
		Standard DenseNet &	92.02 & 99.55 &	99.09\\
		Standard CapsNet & 81.31 & 86.21 & 95.45\\
		Regularized CapsNet & 85.05 & 96.67 & 98.79\\
	    GMP-CapsNet (ours) & \textbf{98.59} & \textbf{99.85} & \textbf{100}\\
		\Xhline{2\arrayrulewidth}

	\end{tabular}

\end{table}

\begin{table}[h]
	\caption{Evaluation results of different methods on MCYT dataset with three protocols 25-75, 50-50, and 25-75\% ratio of training and test data for each signer.}	
	\label{tab:table 4}				
	\centering
	\begin{tabular}{ l c c c}
	    
	    \Xhline{2\arrayrulewidth}

	     Methods & \multicolumn{3}{c}{MCYT}\\
	    \cline{2-4}
	    	& 25-75 & 50-50 & 75-25\\
		\hline
		Standard DenseNet &	 85.21 	& 95.05 & 	 98.33 \\
		Standard CapsNet &  68.85  & 80.76 & 87.00 \\
		Regularized CapsNet &  76.85 & 92.00 & 98.33 \\
	    GMP-CapsNet (ours) & \textbf{94.55} & \textbf{99.24} & \textbf{99.66}\\
		\Xhline{2\arrayrulewidth}

	\end{tabular}	
\end{table}

\begin{table}[h]
	\caption{Evaluation results of different methods on the UTSig dataset with three protocols 25-75, 50-50,  and 25-75\% ratio of training and test data for each signer}	
	\label{tab:table 5}				
	\centering
	\begin{tabular}{ l c c c}
	    
	    \Xhline{2\arrayrulewidth}

	     Methods & \multicolumn{3}{c}{UTSig}\\
	    \cline{2-4}
	    	& 25-75 & 50-50 & 75-25\\
		\hline
		Standard DenseNet &	 86.04 	& 96.99 & 98.63  \\
		Standard CapsNet &  74.57 & 84.35 & 87.58  \\
		Regularized CapsNet &  85.13 & 93.85 & 96.27  \\
	    GMP-CapsNet (ours) & \textbf{94.30} & \textbf{98.19} & \textbf{99.38}\\
		\Xhline{2\arrayrulewidth}

	\end{tabular}	
 
\end{table}

\subsection{Comparisons to state-of-the-art signature recognition}
Several studies in signature recognition on the datasets mentioned above, such as \cite{calik2019large,jampour2019chaos,gumusbas2019offline,hadjadji2017efficient,djoudjai2017offline,boudamous2017open}, are considered in this section. To this end, we follow the same experimental settings. One of the successful and recent works in signature recognition is presented in \cite{calik2019large}. This study implements multiple deep methods, including standard VGG-16 and two parametric versions of VGG-S and VGG-M. As shown in Table \ref{tab:table 6}, except for one protocol, the proposed method obtained the best results in all experiments. In addition, a study on similar datasets based on the capsule networks is presented in \cite{gumusbas2019offline}, which significantly improved compared to classical approaches. The results of this method are also given in the comparisons. The details of these results are shown in Table \ref{tab:table 6}.

\begin{table}[h]
	\caption{Evaluation comparisons of the proposed method with state-of-the-art on two datasets, CEDAR and MCYT}	
	\label{tab:table 6}				
	\centering
	
	\begin{tabular}{ l c c c | c c c}
	    
	    \Xhline{2\arrayrulewidth}

	     Methods & \multicolumn{3}{c}{CEDAR} & \multicolumn{3}{c}{MCYT}\\
	    \cline{2-7}
	    	& 25-75 & 50-50 & 75-25 & 25-75 & 50-50 & 75-25\\
		\hline
		VGG-16 \cite{calik2019large} & 96.59 & 98.15 & NA & 90.95 & 95.93 & NA\\ 
VGG-S \cite{calik2019large} & 97.13 & 98.15 &  NA & 93.47 & 97.53  &  NA \\
VGG-M \cite{calik2019large} & 97.35 & 98.75 & NA &  93.96 & 97.73  & NA \\
LS2Net \cite{calik2019large} & 98.30 & 98.88 &  NA & \textbf{96.41} & 98.13 & NA \\
OffCapsNet \cite{gumusbas2019offline}  &  NA & 97.20 & 98.80 &  NA & NA & NA \\
Standard DenseNet & 92.02 & 99.55 & 99.09 & 85.21 &  95.05  & 98.33 \\ 
Standard CapsNet & 81.31 & 86.21 & 95.45 &  68.85 & 80.76 & 87.00 \\ 
Regularized CapsNet & 85.05 & 96.67 & 98.79 & 76.85 & 92.00 & 98.33 \\
GMP-CapsNet (ours) & \textbf{98.59} & \textbf{99.85
} & \textbf{100} & 94.55 & \textbf{99.24} & \textbf{99.66} \\

		\Xhline{2\arrayrulewidth}

	\end{tabular}	
 
\end{table}

In another setting, evaluations have been performed on the CEDAR dataset using only five signature images for training and 19 similar test images \cite{jampour2019chaos,hadjadji2017efficient,djoudjai2017offline,boudamous2017open}. The results are presented in Table \ref{tab:table 7}, in which the proposed method performs better than other related works.

\begin{table}[h]				
	\caption{The results of the proposed method compared to other studies on the CEDAR dataset with only five training data for each class to measure generalizability}
	\label{tab:table 7}				
	\centering
	
	\begin{tabular}{ l l c }
	    \Xhline{2\arrayrulewidth}
	    Methods	& Features & Accuracy(\%)\\
		\hline
  CT-es \cite{hadjadji2017efficient} & Curvelet Transform (Equi-Spaced) & 92.05 \\
HOT \cite{boudamous2017open} & Histogram Of Templates & 92.50\\
CG \cite{chang2020ai} & Chaos Game with Fractal theory & 97.43\\
CT-em \cite{hadjadji2017efficient} & Curvelet Transform (Equi-Mass) & 97.51\\
CT-FI \cite{hadjadji2017efficient} & Curvelet Transform (Fuzzy Integral) & 97.99\\
HSR \cite{hadjadji2017efficient} & Histogram of Symbolic Represent &	98.63 \\
GMP-CapsNet (ours) & Generalized Multi-prototype CapsNet	& \textbf{98.76}\\
		\Xhline{2\arrayrulewidth}
	\end{tabular}	
 
\end{table}

A complementary experiment with similar settings is also presented in \cite{foroozandeh2020offline}, \cite{foroozandeh2020use} on the UTSig dataset to make a fair comparison with state-of-the-art studies. In these settings, the first 60\% of signatures are used for training and the last 20\% for testing. As shown in Table \ref{tab:table 8}, the proposed method has performed better than competitors on this dataset, so that we can see a significant gap in some results.

\begin{table}[h]				
	\caption{The results of the proposed method compared to the state-of-the-art on the UTSig dataset}
	\label{tab:table 8}					
	\centering
	
	\begin{tabular}{ l l c }
	    \Xhline{2\arrayrulewidth}
	    Methods	& Features & Accuracy(\%)\\
		\hline
 CG \cite{jampour2019chaos} & Chaos Game with Fractal theory & 71.48\\ Conv.Net & Personal Conv.Net & 96.29 \\
 CBR	& Global and local feature values & 97.19\\
 Shearlet \cite{jampour2019chaos} & Transfer Learning with Shearlet Transform & 95.24 \\
ResNet \cite{jampour2019chaos} & ResNet-50 & 92.59 \\
Inception & Inception-V3 & 97.58 \\
VGG-19 \cite{jampour2019chaos} & VGG-19 & 98.55 \\
VGG-16 \cite{jampour2019chaos} & VGG-16 &98.71 \\
GMP-CapsNet (ours) & Generalized multi-prototype capsule network	& \textbf{99.13} \\
		\Xhline{2\arrayrulewidth}
	\end{tabular}	
 
\end{table}

\subsection{Statistical analysis}
In order to statistically analyze and prove the significance of the difference between the results of the proposed method and other methods, we used a Wilcoxon signed-rank test. The analysis of the classification results in Table \ref{tab:table 2} between the proposed method and the standard capsule network method indicates a significant difference, with 97.23\% confidence. Also, considering the results in Table \ref{tab:table 3}, Table \ref{tab:table 4}, and Table \ref{tab:table 5}, the difference between the Standard DenseNet, Standard CapsNet, and Regularized CapsNet models compared to the proposed GMP-CapsNet model is 99.23\% significant. Accordingly, the proposed method achieves promising performance compared to other methods.

\subsection{Part-level multi-prototyping analysis}
In \cite{ghiasi2019competitive} and \cite{abbaasi2023multi}, it has been shown that neurons and capsules of the final layer represent different intra-class clusters, respectively. In this section, we analyze the multi-prototyping performance in the middle layers of the capsule network. Therefore, consider a multi-prototype capsule neural network with three capsule layers. This section aims to visualize the prototypes discovered by the middle capsule layer. We expect to see different part prototypes in the middle layer.

\begin{figure}[htp]				
	\centering
	\includegraphics[scale=.3]{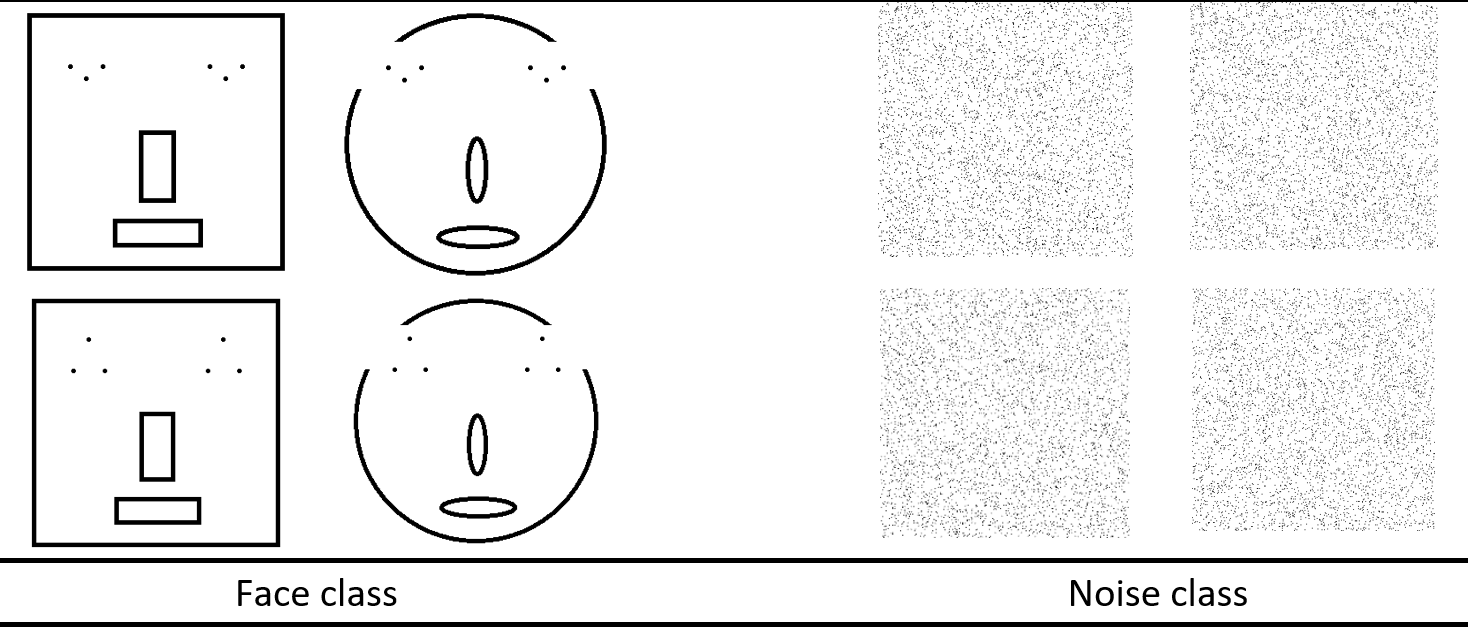}
	\caption{Synthetic face and noise dataset. The eye prototypes and other parts are depicted in the face class.}
	\label{fig:figure 3}
\end{figure}

We could not perform this analysis on a complex real-world dataset since the high complexity of real images makes the network choose particular parts to identify classes that are probably not reasonable to humans for the recognition task. Therefore, we designed a toy dataset so that the network has no choice but to identify predetermined target parts. Some examples of the toy dataset are illustrated in Figure \ref{fig:figure 3}. The goal is to classify toy faces from a set of noise images. As it is clear, in the face class, the eyes have two prototypes shown in different rows. Each prototype consists of sub-parts. The eye prototypes are randomly shifted by a maximum of 3 pixels and randomly scaled by a ratio between $0.7$ and $1.0$.

\begin{figure}[htp]				
	\centering
	\includegraphics[scale=.3]{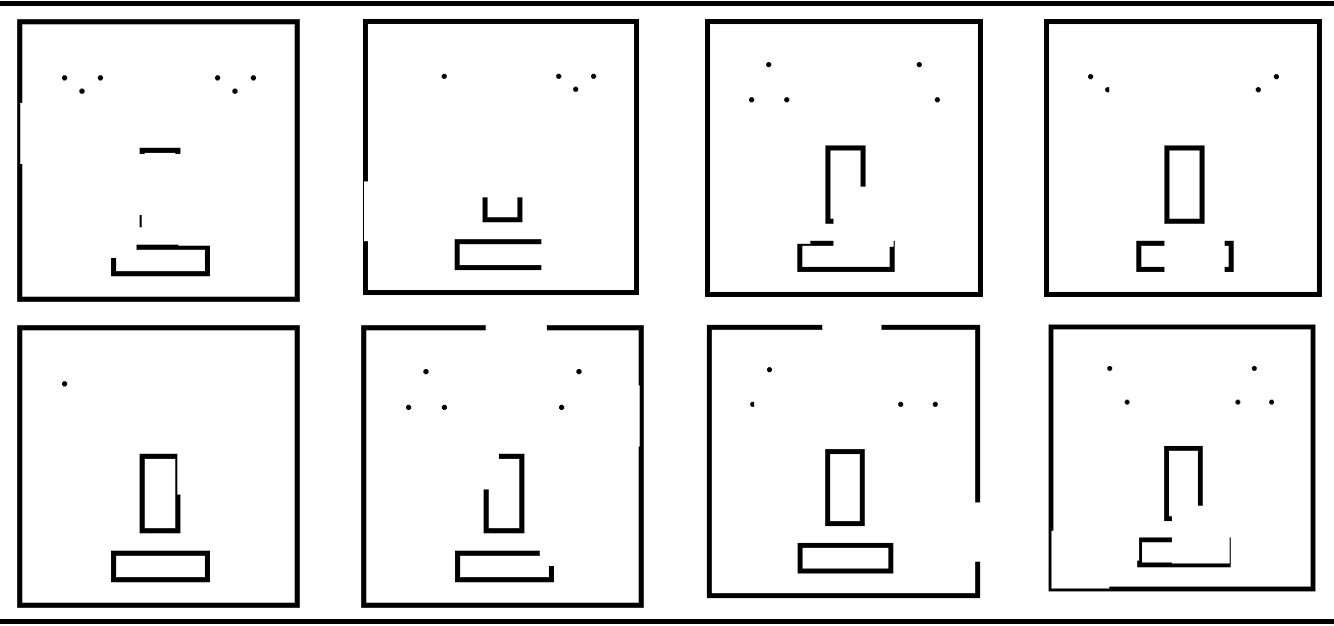}
	\caption{Illustration of some face samples after dropout to prevent the network from focusing on a specific part of the image}
	\label{fig:figure 4}
\end{figure}

Since solving such a classification problem is a simple task for the capsule networks, the network will not tend towards identifying different prototypes to achieve maximum accuracy, and recognition is done by representing a few parts of face images. Therefore, in addition to removing the circular emoticon, we perform a dropout in the face class images to prevent the network from focusing on a specific image part. Hence, In this setup, each $64\times 64$ image is randomly manipulated by five $10\times 10$ patches. Figure \ref{fig:figure 4} illustrates some face samples after dropout.

\begin{figure}[htp]				
	\centering
	\includegraphics[scale=.3]{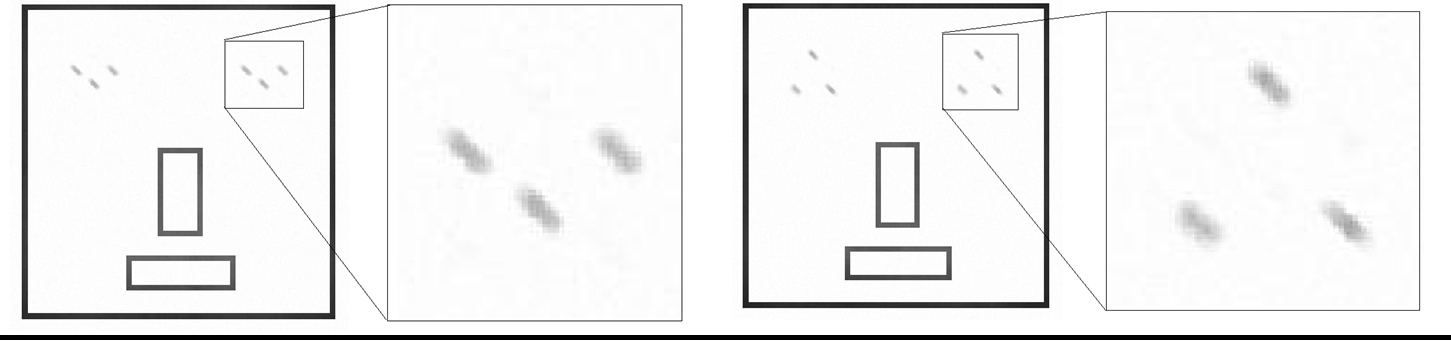}
	\caption{The results of the average eye prototype for the face class images in each of the two identified prototypes. According to these results, the multi-prototype network tends to learn different image part prototypes in the middle layer of the multi-prototype}
	\label{fig:figure 5}
\end{figure}

We sampled 600 images for each class. In the first capsule layer, multi-prototyping has not been carried out. The reason is that we want to force the network to identify different eyes only in the middle layer. In the same way, in the final capsule layer, multi-prototyping has not been done in order not to discover two different face prototypes. Therefore, in this setup, only the middle layer of the capsule is equipped with multi-prototyping. Considering the large number of capsules in the middle layer, the detection of capsules related to eye parts is done by supplying an extra label for two different eye prototypes for images. The results of the average eye area for images activated by each of the two prototypes can be seen in Figure \ref{fig:figure 5}. The obtained results show the expected performance of the network in identifying different part prototypes in the middle layer of the multi-prototype capsule network.

\section{Conclusion}
\label{sec:conclusion}
Capsule neural networks identify image parts and form ``whole'' instantiation parameters based on parts. Since a high intra-class variation is generally presented in real-world datasets, considering one capsule per ``object'' or ``whole'' is insufficient to model the complexity of a ``whole'' well. In this paper, we proposed a multi-prototype architecture by which the network tends to learn the diversity in the dataset. The proposed method extends our previous study in \cite{abbaasi2023multi}. We adopt several capsules instead of considering a single capsule for each class and each part. Co-group capsules compete during the learning process, and for each image, the winner regards as the output of each group. Next, to improve the capsule network input, an architecture is utilized in which the input of the capsule network is connected to the middle layers of the DenseNet, which makes the network perform better because of richer and more robust features for image affine transformation. Because in multi-prototyping at the part level, there is a possibility of part aligning problems. The experiments conducted in image classification show the superiority of the proposed model compared to the standard capsule network and other studies in terms of classification accuracy. We also empirically showed that the middle layers prototypes capture the intra-part varieties in the proposed architecture.


\end{document}